\documentclass{article}

\usepackage{arxiv}
\usepackage[square,numbers]{natbib}
\usepackage{multicol}
\usepackage{url}
\usepackage{gb4e}
\noautomath

\usepackage{fancyhdr}
\fancypagestyle{plain}{%
  \fancyhf{} 
  \fancyfoot[C]{\tiny Copyright \textcopyright\ 2009 All rights reserved. Duplication for purposes of any kind is strictly
    forbidden.}
  
}

\usepackage{leipzig}

\newleipzig{aor}{aor}{aorist}
\newleipzig{attr}{attr}{attributive}

\usepackage{listings}
\usepackage{multirow}
\usepackage{tikz-dependency}

\makeatletter
\newcommand\footnoteref[1]{\protected@xdef\@thefnmark{\ref{#1}}\@footnotemark}
\makeatother

\date{}
\title{Enhancements to the BOUN Treebank Reflecting the Agglutinative Nature of Turkish}

\author{Büşra Marşan\textsuperscript{1}  \\\texttt{busra.marsan@boun.edu.tr} \and
Salih Furkan Akkurt\textsuperscript{2}\\\texttt{furkan.akkurt@boun.edu.tr} \and
Muhammet Şen\textsuperscript{2}\\\texttt{muhammet.sen@boun.edu.tr} \and
Merve Gürbüz\textsuperscript{2}\\\texttt{merve.gurbuz@boun.edu.tr} \and
Onur Güngör\textsuperscript{2}\\\texttt{onurgu@boun.edu.tr} \and
Şaziye Betül Özateş\textsuperscript{2}\\\texttt{saziye.bilgin@boun.edu.tr} \and
Suzan Üsküdarlı\textsuperscript{2}\\\texttt{suzan.uskudarli@boun.edu.tr} \and
Arzucan Özgür\textsuperscript{2}\\\texttt{arzucan.ozgur@boun.edu.tr} \and
Tunga Güngör\textsuperscript{2}\\\texttt{gungort@boun.edu.tr} \and
Balkız Öztürk\textsuperscript{1}\\\texttt{balkiz.ozturk@boun.edu.tr}\\[6pt]
1. Linguistics Department, Boğaziçi University\\
2. Computer Engineering Department, Boğaziçi University
}

\begin{document}
\maketitle
\begin{abstract}
  In this study, we aim to offer linguistically motivated solutions to resolve the issues of the lack of representation of null morphemes, highly productive derivational processes, and syncretic morphemes of Turkish in the BOUN Treebank without diverging from the Universal Dependencies framework.  
  In order to tackle these issues, new annotation conventions were introduced by splitting certain lemmas and employing the MISC (miscellaneous) tab in the UD framework to denote derivation. Representational capabilities of the re-annotated treebank were tested on a LSTM-based dependency parser and an updated version of the BoAT Tool is introduced.
 
\end{abstract}


\section{Introduction}
\label{intro}
Following the dependency grammar framework first proposed by Tesniére \cite{tesniere}, dependency trees illustrate how sentence elements relate to one another through head and dependent relations. Universal Dependencies\footnote{\url{https://universaldependencies.org}} (UD) is an international cooperative treebank project based on the dependency grammar framework and it aims to offer a standardized and comprehensive dependency treebank collection covering 121 languages.

With the addition of new UD treebanks, Turkish does not qualify as a low resource language anymore. With a total of 733,000 tokens, it is the 12th largest UD treebank in the UD repository. Although the coverage of the treebanks plays an essential role in improving the performance of natural language processing (NLP) systems \cite{foth2014because}, their ability to correctly and consistently illustrate the morphosyntactic features of the target language should not be overlooked. As Vincze \textit{et al.} \cite{vincze2017universal}’s study shows, the better a treebank’s ability to represent the morphology and syntax of the target language, the better the performance of the NLP systems using that treebank as a resource. 

In this paper, we aim to abide by the linguistic framework set by Bedir \textit{et al.} \cite{bedir2021overcoming} and offer an updated and comprehensive UD treebank for Turkish, the BOUN Treebank, along with an improved UD annotation interface, the BoAT Tool, first introduced in Türk \textit{et al.} \cite{turk2021resources}.

The decisions made in the re-annotation process of the BOUN Treebank aim to offer solutions to the issues posed by the morphologically rich and complex nature of Turkish: null morphemes are frequently employed, agglutinative processes are heavily used to create new forms, and numerous morphemes like copula and -ki are very syncretic. The main goal of this study is to illustrate these phenomena without compromising the compliance with the UD framework.  

This paper is organized as follows. Previous attempts at creating dependency treebanks in Turkish are laid out in Section~\ref{litrev}. Annotation changes made in the current version of the BOUN Treebank and their linguistic justification are discussed in Section~\ref{annotation}. Statistics regarding the changes made to the previous version are stated in Section~\ref{stats}. Improvements made to the BoAT Annotation Tool are explained in Section~\ref{boat}. Finally, the performance of the parser trained using the current version of the treebank is reviewed in Section~\ref{parser}.

\section{Dependency Treebanks in Turkish}
\label{litrev}
Shortly after the first dependency treebank for Turkish was presented by Atalay \textit{et al.} \cite{atalay2003annotation}, Eryiğit and Pamay \cite{eryiugit2007itu} offered a smaller dependency treebank consisting of 300 sentences as part of the CoNLL 2007 Shared Task: MST Treebank. 13 years after its publication, Sulubacak et al. \cite{sulubacak2016universal} re-annotated this dataset, converted it to the UD framework, and published the updated dataset as IMST-UD. 
Çöltekin’s The Grammar Book Treebank (GB) \cite{coltekin2015grammar} which consists of 2,803 sentences extracted by a reference book on Turkish grammar by Göksel and Kerslake \cite{goksel2004turkish} marks the very first effort in creating the first UD-style Turkish treebank.  Another pioneer in Turkish dependency treebanks is IWT-UD as it is the first Turkish dependency treebank that covers informal texts. IWT-UD was introduced by Sulubacak and Eryiğit \cite{sulubacak2018implementing} three years after a constituency-style treebank, IWT, was presented by Pamay et al. \cite{pamay2015annotation}. 

Tourism is one of the two domain-specific dependency treebanks in Turkish and consists of hotel and restaurant reviews. The other one is ATIS treebank that covers the Turkish translation of English ATIS (Airline Travel Information System) corpus \cite{hemphill1990atis}. 

Other Turkish UD-style treebanks include Kenet UD Treebank, Penn Treebank, and FrameNet treebank. Penn Treebank consists of Turkish translations of English Penn Treebank \cite{taylor2003penn} while FrameNet consists of 2,700 sentences from the Turkish FrameNet database \cite{marcsan2021building}. 

With 9,761 sentences and 121,214 tokens randomly selected from Turkish National Corpus (TNC) \cite{aksan2012construction}, the BOUN Treebank is one of the largest UD-style treebanks in Turkish. Covering five different registers (broadsheet national newspapers, biographical texts, essays, popular culture articles, and instructional texts), it offers word order and sentence length variance in addition to linguistically motivated dependency annotations (see Section~\ref{annotation}).

\section{Improving BOUN Treebank}
\label{annotation}
The first step of the previous annotation process of the BOUN Treebank (see \cite{turk2021resources} for a detailed discussion) was parsing the raw text to create CoNLL-U files using Kanerva et al.’s \cite{kanerva2018turku} pipeline tool. During this parsing process, UPOS tags and certain morphological information were automatically annotated. Then the dependency relations were manually annotated by two native speakers of Turkish who are linguists. To ensure inter-annotator agreement, randomly selected 1000 sentences were double annotated. Using Cohen’s Kappa measure, inter-annotator agreement was calculated. Dependency label match score was 0.82, unlabelled attachment score was 0.83, and labelled attachment score was 0.75.

\subsection{The Re-annotation Process: Overcoming the Challenges }
For the re-annotation process, a team of linguists detected shortcomings and problematic annotations of the previous version of the BOUN Treebank \cite{turk2021resources}. Two major representation challenges were detected: derivation and null copula. The following subsections discuss the strategies employed to overcome these challenges.

Annotations were done by two linguists who are native speakers of Turkish. In order to ensure inter-annotator agreement, 100 sentences were double annotated. Unlabelled attachment and labelled attachment scores were calculated using Cohen’s Kappa measure. Their respective values are 98.61 and 97.81.

\subsubsection{Derivation}
\begin{figure}[h!]
    \centering
    \includegraphics[width=5.5cm, height=4cm]{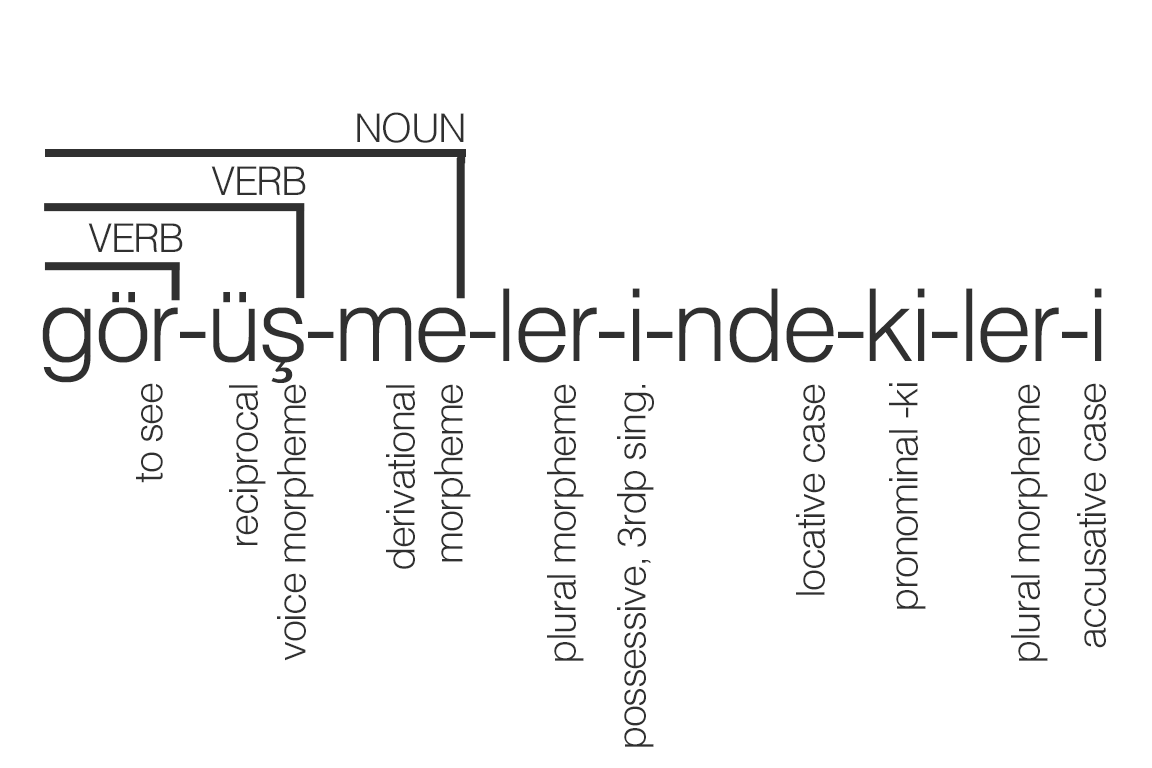}
    \caption{Derivational and inflectional analysis of ``görüşmelerindeki'' (\textit{``those who were/are at his/her meetings''})}
    \label{fig1}
\end{figure}

Having its focus on syntax, the UD framework falls short of representing derivational processes. The official guide of UD argues that the final derivational suffix in a word is opaque in the sense that it does not permit access to the morphemes that come before it. Hence in a construction as the one shown in Figure~\ref{fig1}, numerous derivational and inflectional morphemes before the last derivational morpheme (-ki) are lost.

As one of the primary concerns of this study was finding ways to illustrate derivation processes like those in Figure~\ref{fig1} without diverging from the UD framework, two strategies were employed for different cases:

\begin{itemize}
\item For morphemes like -lI (\textit{``with''}), \texttt{df=} function is introduced in the MISC tab.
\item Lemmas containing -ki morpheme are splitted. 
\end{itemize}

These derivational morphemes imply that the host lemma and its modifier form a syntactic unit together. As a result, the correct bracketing of the -lI adjective, its modifier, and the constituent they modify together should look like (\ref{kedi2}) instead of (\ref{kedi}). Representing the derivational morphemes like -lI and -sIz allows keeping this crucial syntactic information.\footnote{Abbreviations: attr = attributive}

\begin{multicols}{2}

\begin{exe}
\ex \label{kedi}
\gll [ [ [kahverengi] tüy-lü] kedi] \\
{\ } {\ } \phantom{[}brown fur-\Attr{} cat\\
\glt ``a cat with brown fur''
\ex \label{kedi2}
\gll [ [kahverengi tüy]-lü kedi] \\
{\ } \phantom{[}brown fur-\Attr{} cat\\
\glt ``a cat with brown fur''
\end{exe}

\end{multicols}

With the new \texttt{df=} function proposed in this study, ``tüylü'' is not decomposed as two lemmas: ``tüy'' (\textit{``fur''}) and ``lü'' (\textit{``with''}). Instead, it is left intact and \texttt{df=tüy} (``derived from=fur'') function is annotated in the MISC tab.

Another challenge regarding the derivational processes is posed by -ki. There are two different -ki morphemes in Turkish \cite{hankamer2004there}. One is used as noun and the other derives adjectives from nouns.

Statistically, the vast majority of inflectional suffixes in Turkish occur after the derivational ones. However, both types of the -ki morphemes diverge from this distribution as it can be observed in Figure~\ref{fig1}, where the inflectional suffix -ler precedes the derivational -ki. Hence allowing -ki to ``block access'' \cite{bedir2021overcoming}\footnote{It is stated in the UD guidelines that ``the lemma does not remove derivational morphology, so the lemma of [en] ``organizations'' is ``organization'' not ``organize'' (nor ``organ'').'' See \url{https://universaldependencies.org/u/overview/morphology.html} for the full picture.} to the morphemes that were attached before it reduces the capabilities of the annotation to represent a great deal of morphosyntactic information. With the aim of offering a solution to this issue, a decision to split lemmas that contain either type of -ki was made.\footnote{Refer to the Appendix to compare previous and updated annotation schemes for both -ki morphemes.} Although splitting lemmas is not a common practice within the UD framework, we believe that the theoretical motivation behind this decision justifies the divergence from the framework. (For a detailed linguistic discussion of this issue, please refer to Bedir \textit{et al. }\cite{bedir2021overcoming})

\subsubsection{Null Morpheme}
Languages such as Turkish, Russian, Arabic, and Coptic have null morphemes, however, the UD framework does not officially support such phenomena. As a result, independent strategies have emerged to represent null morphemes in these languages. For example, Coptic avoids null subject nodes by using fused forms\cite{zeldes2018coptic}, Marathi annotates the feature values of the null morphemes, however, does not introduce any information (i.e. annotation) to indicate that they are null morphemes\cite{ravishankar2017universal}. Widely employing null morpheme for pluralization, Arabic distinguishes between two types of annotation: Form-based and function-based. In their UD-style dependency treebank annotation, Marton \textit{et al.}\cite{marton2011improving} follow a function-based annotation framework and annotate the feature values of null morphemes.

In Turkish, the copula can surface in three different forms \cite{goksel2004turkish}: i-, -y-, and $\emptyset$.After considering UD guidelines and particularities of Turkish copula, it was decided to employ the MISC tab again by introducing two new functions for the null copula: \texttt{nullcop=3s} (singular) and \texttt{nullcop=3p} (plural).
By following a function-based annotation schema, we were able to offer more linguistically accurate annotations without diverging from the UD framework.

\subsubsection{Copula}
In Turkish, ol- copula has six distinct functions \cite{bedir2021overcoming}: An intransitive verb meaning ``to be suitable/fit'', a transitive verb meaning ``to become'', an auxiliary verb in embedded sentences, an auxiliary verb following the participle, a light verb forming complex verbal constructions (such as ``sorun olmak'' (\textit{``to be/become an issue''})), and finally the existential predicate that surfaces as ``var'' (\textit{``to exist''}) and ``yok'' (\textit{``not to exist''}). Yet the previous annotation scheme of the BOUN Treebank made no distinction between these different usages. To offer more accurate representations, certain annotation changes were made regarding these functions.

\subsection{Newly Introduced XPOS Tags and Dependency Relations}
In an attempt to overcome the challenges thoroughly discussed in the previous subsection, a set of new XPOS tags and dependency relations were introduced in the updated version of the BOUN Treebank. A comprehensive list can be found in Table~\ref{newthings}.

\begin{table}[h!]
\scalebox{0.75}{
\begin{tabular}{l|llllllll}
\textbf{Lemma}  & \textbf{var} & \textbf{yok} & \textbf{\begin{tabular}[c]{@{}l@{}}ol- (after\\ participle)\end{tabular}} & \textbf{\begin{tabular}[c]{@{}l@{}}ol- (in\\ embedded\\ sentences)\end{tabular}} & \textbf{\begin{tabular}[c]{@{}l@{}}ol- (in light verb\\ constructions)\end{tabular}} & \textbf{\begin{tabular}[c]{@{}l@{}}ol- (as transitive\\ or intransitive\\ verb)\end{tabular}} & \textbf{\begin{tabular}[c]{@{}l@{}}-ki\\ \ (adjectivizer)\end{tabular}} & \textbf{\begin{tabular}[c]{@{}l@{}}-ki\\ (pronominal)\end{tabular}} \\ \hline
\textbf{UPOS}   & NOUN         & NOUN         & AUX                                                                      & AUX                                                                             & VERB                                                                                & VERB                                                                                         & PART                                                                  & PRON                                                                \\
\textbf{XPOS}   & Exist        & Exist        &                                                                          & Ptcp                                                                            &                                                                                     &                                                                                              & Attr                                                                  & Partic                                                              \\
\textbf{Deprel} & root         & root         & aux                                                                      & cop                                                                             & compound:lvc                                                                        & root                                                                                         & dep:der                                                               &                                                                    
\end{tabular} }

\caption{\label{newthings}New dependency relations and XPOS tags proposed for the updated BOUN Treebank.}
\end{table}

\section{Statistics}
\label{stats}
As part of the re-annotation process, 117,732 changes were made in the following tabs: UPOS, XPOS, Deprel, MISC, and Features. The majority of annotation changes targeted UPOS and XPOS tags (see Table~\ref{changes1}). Since morphological information was automatically annotated in the previous version of the treebank by the parsing tool \cite{kanerva2018turku}, refinements by the annotators were required in order to ensure accuracy.

\begin{table}[h!]
\begin{tabular}{l|lllll}
\textbf{Field}   & \textbf{UPOS} & \textbf{XPOS} & \textbf{Features} & \textbf{Deprel} & \textbf{MISC} \\ \hline
\textbf{Changes} & 11,396         & 63,829         & 27,098             & 23,32            & 4,973         
\end{tabular}
\caption{\label{changes1}Changes made in the re-annotation process.}
\end{table}

The changes in UPOS values reflect the linguistics-based decisions made in the re-annotation process. Previous version of BOUN Treebank made no distinction between two -ki morphemes in Turkish. As a result, almost all -ki instances were labeled as \texttt{CConj} (clausal conjunction). After deciding to make a distinction between adjectivizer -ki and pronominal -ki, the UPOS tag of the former was changed to \texttt{Part}. 

\begin{table}[h!]
\scalebox{0.90}{
\begin{tabular}{c|ccc|ccc}
\textbf{Field}  & \multicolumn{3}{c|}{\textbf{UPOS}}                                                                      & \multicolumn{3}{c}{XPOS}                                                                                 \\ \hline
\textbf{Change} & \textbf{Adj -\textgreater Noun} & \textbf{CConj -\textgreater Part} & \textbf{Noun -\textgreater Propn} & \textbf{Verb -\textgreater Ptcp} & \textbf{Verb -\textgreater Vnoun} & \textbf{ANum -\textgreater Indef} \\ \hline
\textbf{Count}  & 1,595                            & 1,025                              & 968                               & 2,459                             & 1,664                              & 1,622                             
\end{tabular} }
\caption{\label{changes2}Most frequent changes targeting the UPOS and XPOS tags}
\end{table}

Due to the shortcomings of automatic morphological tagging, some proper nouns were labeled as nouns. Re-annotation process targeted them as well: UPOS tags of 986 proper nouns were changed. In addition, XPOS tags of verbal nouns and participle forms were updated (see Table~\ref{changes2}).

\section{The BoAT Tool}
\label{boat}
The BoAT tool offered in Türk et al. \cite{turk2021resources} is a desktop application for manually annotating sentences parsed by a dependency parser. In the scope of the current work, in addition to the reannotation of the BOUN Treebank, we enhanced the tool with additional functionalities. We will publish the tool as open source on GitLab accompanied with a user manual.

\subsection{Changes}
BoAT is a desktop application, written in Python and based on Qt. The Qt version has been incremented from 5 to 6. This resulted in a more modern-looking user interface (UI). With ample feedback from the annotators who had used the tool for the BOUN Treebank, some requested features and improvements have been surfaced.

\textit{Clutter}: Annotation table's columns were being shown or hidden by checkboxes above the table. These were removed and a textbox beside the other buttons has been added to replace them. This textbox serves exactly the same purpose while taking less space.

\textit{Autocompletion}: Annotators use the tool for long hours at a time. Thus after a while, mistakes tend to occur. Another requested feature, autocompletion of the table fields, aims to prevent such mistakes. Many fields of the table have predetermined sets of values they can take. By not allowing values outside these sets and having a shorthand writing system whereby an annotator enters only the start of a value and it gets filled automatically, we implemented this much requested feature.

\textit{Saving as CoNLL-U}: Another change was regarding saving of the \textit{CoNLL-U} documents. The initial tool saved every edit automatically, yet treebanks with thousands of sentences tend to take time to save. This mishap seemed to slow our annotators. We added a save button instead of the autosave feature. Currently, a user uses the save button to edit the actual \textit{CoNLL-U} file.

\textit{Shortcuts}: The initial tool already had shortcut support. Going with the focus-oriented approach, all the new tasks have keyboard shortcuts associated with them as well. This alleviates the need to use a mouse while annotating via keyboard.

\textit{Dependency graphs}: The vertical dependency graph in the initial version was replaced by the horizontal dependency graph of spaCy \cite{spacy}, displaCy, due to the spatial concerns.

\textit{Validation}: The validation script for annotations has been upgraded to the latest version written by the UD framework, which has much more detailed explanations for why a specific annotation is invalid.

\section{Parser Performance}
\label{parser}
Within the scope of this study, an NLP task was conducted. BiLSTM-based biaffine dependency parser proposed by Dozat and Manning \cite{dozat2017stanford} was trained using the updated BOUN Treebank. The train set contained 7.803 sentences, development set contained 982 sentences and test set contained 979 sentences. Considering the average arc length and average token count (see Table~\ref{dataset} for details) of each set, BOUN Treebank offers a well-balanced data set.

\begin{table}[h!]
\begin{tabular}{l|l|l|l|l}
                                                                       & \textbf{Train} & \textbf{Development} & \textbf{Test} & \textbf{Entire Data} \\ \hline
\textbf{\begin{tabular}[c]{@{}l@{}}Average\\ Arc Length\end{tabular}}  & 2.91           & 2.88                 & 2.82          & 2.90                 \\ \hline
\textbf{\begin{tabular}[c]{@{}l@{}}Average\\ Token Count\end{tabular}} & 12.83          & 12.42                & 12.36         & 12.74                \\ \hline
\textbf{\begin{tabular}[c]{@{}l@{}}Number\\ of Sentences\end{tabular}} & 7,803          & 982                  & 979           & 9,761               
\end{tabular}
\caption{\label{dataset}Sizes and specifications of train, development and test data sets}
\end{table}

The previous version of the BOUN Treebank \cite{turk2021resources} yielded 77.36 unlabeled attachment score (UAS) and 70.37 labeled attachment score (LAS). After being trained on the new dataset, UAS is increased by 0.59 points to reach 77.96 while LAS showed a 0.10 decrease by dropping to 70.26 points. After the re-annotation process, several faulty or disputed dependency relations were fixed in the data, hence a rise in UAS was observed.

In order to better account for the particularities of Turkish morphosyntax, four new dependency labels were introduced: \texttt{dep:der} for adjectivizer -ki, \texttt{obl:tmod} for obliques that offer temporal information regarding the predicate, \texttt{advmod:emph} for dA clitics, and \texttt{compound:lvc} for light verb constructions with ol- copula. Moreover, new use cases for \texttt{cop} dependency label were offered to represent different functions of the ol- copula. These changes in the annotation framework added four new dependency types, thus the total class number was increased. 

A sum of 1,032 \texttt{dep:der}, 894 \texttt{obl:tmod}, 1,860 \texttt{advmod:emph}, and 1,545 \texttt{compound:lvc} tags were added. Newly introduced tags and classes introduced added complexity, hence they might be the reason behind the slight decrease in the LAS results. 

The main annotation changes made as part of this study were focused on morphology yet BiLSTM-based biaffine dependency parser proposed by Dozat and Manning \cite{dozat2017stanford} doesn't refer to the morphological information. In fact, it almost completely ignores the morphology features while parsing. Hence the significance of the improvements offered by this study can be better gauged by a morphology-aware parser or a morphological analysis based downstream task.

\section{Conclusion and Further Research}
\label{conc}
The UD framework highly emphasizes syntactic relations and aims to offer a universal foundation to represent typologically different languages in a uniform way. While doing so, certain particularities of these languages tend to get lost in the annotation process in an attempt to abide by the UD convention. The aim of this study is to offer linguistically sound solutions to illustrate syntactically relevant morphological features of Turkish such as null morpheme realizations and derivational processes without diverging significantly from the UD framework.

In order to test the morphological capabilities of the re-annotated BOUN Treebank, a parser that refers to morphological information can be implemented in further research.

\textbf{acknowledgments}
This work was supported by Boğaziçi University Research Fund Grant Number 16909. TUBAGEBIP Award of the Turkish Science Academy (to A.O.) is gratefully acknowledged.
\bibliography{main}

\appendix
\section{A comparison of previous and updated annotation schemes}
\begin{table}[h]
\scalebox{0.87}{
\begin{tabular}{lllllllll}
\hline
Token & Form      & Lemma     & UPOS & XPOS & Features                                                                                                                           & Head & Deprel & MISC \\ \hline
1     & başındaki & başındaki & ADJ  & Adj  &                                                                                                                                    & 2    & amod   &      \\
3     & şapkası   & şapka     & NOUN &      & \multirow{2}{*}{\begin{tabular}[c]{@{}l@{}}Case=Nom|Number=Sing|Number{[}psor{]}=\\ Sing|Person=3|Person{[}psor{]}=3\end{tabular}} & 0    & root   &      \\
      &           &           &      &      &                                                                                                                                    &      &        &      \\ \hline
\end{tabular}}
\caption{\label{ki1a}Previous annotation scheme for ``başındaki şapkası'' (\textit{``the hat on his/her head''})}
\end{table}

\begin{table}[h]
\scalebox{0.87}{
\begin{tabular}{lllllllll}
\hline
Token & Form      & Lemma     & UPOS & XPOS & Features                                                                                                                           & Head & Deprel  & MISC \\ \hline
1-2   & başındaki & başındaki &      &      &                                                                                                                                    &      &         &      \\
1     & başında   & baş       & NOUN &      & \multirow{2}{*}{\begin{tabular}[c]{@{}l@{}}Case=Loc|Number=Sing|Number{[}psor{]}=\\ Sing|Person=3|Person{[}psor{]}=3\end{tabular}} & 3    & nmod    &      \\
      &           &           &      &      &                                                                                                                                    &      &         &      \\
2     & ki        & ki        & PART & Attr &                                                                                                                                    & 1    & dep:der &      \\
3     & şapkası   & şapka     & NOUN &      & \multirow{2}{*}{\begin{tabular}[c]{@{}l@{}}Case=Nom|Number=Sing|Number{[}psor{]}=\\ Sing|Person=3|Person{[}psor{]}=3\end{tabular}} & 0    & root    &      \\
&       &       &      &      &                                 &   & &      \\ \hline
\end{tabular}}
\caption{\label{ki1b}Updated annotation scheme for ``başındaki şapkası'' (\textit{``the hat on his/her head''})}
\end{table}

\begin{table}[h]
\scalebox{1}{
\begin{tabular}{lllllllll}
\hline
Token & Form    & Lemma   & UPOS & XPOS & Features             & Head & Deprel & MISC \\ \hline
1     & seninki & seninki & NOUN &      & Case=Nom|Number=Sing & 0    & root   &      \\ \hline
\end{tabular}}
\caption{\label{ki2a}Previous annotation scheme for ``seninki'' (\textit{``that of yours''})}
\end{table}

\begin{table}[h!]
\scalebox{0.87}{
\begin{tabular}{lllllllll}
\hline
Token & Form    & Lemma   & UPOS & XPOS   & Features                                   & Head & Deprel    & MISC \\ \hline
1-2   & seninki & seninki &      &        &                                            &      &           &      \\
1     & senin   & sen     & PRON & PERS   & Case=Gen|Number=Sing|Person=2|PronType=Prs & 2    & nmod:poss &      \\
2     & ki      & ki      & PRON & Partic & Case=Nom|Number=Sing                       & 0    & root      &      \\ \hline
\end{tabular}}
\caption{\label{ki2b}Updated annotation scheme for ``seninki'' (\textit{``that of yours''})}
\end{table}

\end{document}